\documentclass[times,twocolumn,final,authoryear]{elsarticle}

\usepackage{prletters}
\usepackage{framed,multirow}

\usepackage{amsmath,amssymb,amsfonts}
\usepackage{latexsym}

\usepackage{gensymb}
\usepackage{dblfloatfix}
\usepackage[export]{adjustbox}
\usepackage{tikz}

\usepackage{graphicx}
\usepackage{tabularx}
\usepackage{booktabs}
\usepackage{dcolumn}
\usepackage{hyperref}
\usepackage{subcaption}

\usepackage{url}
\usepackage{xcolor}
\definecolor{newcolor}{rgb}{.8,.349,.1}

\journal{Pattern Recognition Letters}

\begin{document}

\ifpreprint
  \setcounter{page}{1}
\else
  \setcounter{page}{1}
\fi

\begin{frontmatter}

\title{Solving the Same-Different Task with Convolutional Neural Networks}

\author[1]{Nicola \snm{Messina}\corref{cor1}} 
\cortext[cor1]{Corresponding author: 
  Tel.: +39-050-6213054;}
\ead{nicola.messina@isti.cnr.it}
\author[1]{Giuseppe \snm{Amato}}
\author[1]{Fabio \snm{Carrara}}
\author[1]{Claudio \snm{Gennaro}}
\author[1]{Fabrizio \snm{Falchi}}

\address[1]{ISTI-CNR, Via Giuseppe Moruzzi 1, Pisa - 56124, Italy}


\begin{abstract}
Deep learning demonstrated major abilities in solving many kinds of different real-world problems in computer vision literature. However, they are still strained by simple reasoning tasks that humans consider easy to solve.
In this work, we probe current state-of-the-art convolutional neural networks on a difficult set of tasks known as the \textit{same-different} problems.
All the problems require the same prerequisite to be solved correctly: understanding if two random shapes inside the same image are the same or not. 
With the experiments carried out in this work, we demonstrate that residual connections, and more generally the skip connections, seem to have only a marginal impact on the learning of the proposed problems. 
In particular, we experiment with DenseNets, and we examine the contribution of residual and recurrent connections in already tested architectures, ResNet-18, and CorNet-S respectively.
Our experiments show that older feed-forward networks, AlexNet and VGG, are almost unable to learn the proposed problems, except in some specific scenarios.
We show that recently introduced architectures can converge even in the cases where the important parts of their architecture are removed.
We finally carry out some zero-shot generalization tests, and we discover that in these scenarios residual and recurrent connections can have a stronger impact on the overall test accuracy.
On four difficult problems from the SVRT dataset, we can reach state-of-the-art results with respect to the previous approaches, obtaining super-human performances on three of the four problems.
\end{abstract}


\end{frontmatter}


\section{Introduction}
\noindent With the advent of deep learning, the computer-vision world gained a huge boost in almost all its fields.
One of the major innovations in this field was the introduction of Convolutional Neural Networks (CNNs)~\citep{LeCun1998}. CNNs constitute now a standard approach to transform a raw matrix of pixels into some higher-level representation, and they are used in many downstream tasks.


%

Despite their success, there are many open problems with current deep architectures, and in particular with their abstract reasoning abilities. They are still unable to distill high-level general concepts that can be transferred to different domains. This brings to low generalization abilities and often to an overfit to the specific domain on which the network is trained. 
Humans can recognize some shape patterns never seen before, and they can deduce some general properties of a never seen shape (e.g., is it a closed shape? is it the same shape as another one but rotated?).


In this work, we probe modern state-of-the-art vision architectures to understand their ability to draw abstract conclusions on the objects contained in images. We tackle the \textit{same-different} tasks, which consist of a set of problems having a common underlying difficulty: predicting if two shapes inside the same image are the same or not.
It is a challenging set of tasks for convolutional architectures since they are required not to learn specific shape patterns to solve the problem; instead, they require to grow some abstract internal representation which is powerful enough to draw a logical conclusion on a fact hidden in the image (e.g., the shapes in the images are the same even if they are orientated in different ways).

Same-different problems are really interesting challenges. 
Humans perceive the world as a complex set of patterns composite together to form higher-level structures, such as the repeating chorus in a song. By tackling the same-different concept we can better understand the abstract abilities of current deep neural network models, even outside of the computer vision world. The long-term results from these studies can be applied in a wide range of disciplines, from robotics to cultural heritage preservation.


This paper is an extension of our previous work~\citep{Messina2019SameDifferent}, where 
we tested a variety of state-of-the-art deep convolutional architectures on some challenging same-different tasks. 
Preliminary experimental results suggested that residual connections could have been the most important architectural detail for solving this task.

In this work, instead, we show through an improved set of experiments that ResNet-like skip connections are probably not the only essential architectural details triggering the convergence. In this regard, we probed an extended set of state-of-the-art models. In particular, we try to remove the key distinctive architectural features from some of the previously tested models, such as residual connections from ResNets or residual and/or recurrent connections from the CorNet-S architectures. 

We also show that the older VGG-19 and AlexNet architectures, in some specific cases, are able to move away from pure chance accuracy on the test set, although they cannot reach state-of-the-art results on the presented problems.

In this work, we also perform final zero-shot generalization tests on the converged architectures, and we show that residual and recurrent connections can have possibly strong impacts on the final test accuracies.

Despite the underlying difficulty in discerning what are the key architectural factors driving the convergence of these networks, we think that this work evidences in a systematic way the current weaknesses of current vision models.


To sum up, in this paper, we extend the previous work~\citep{Messina2019SameDifferent} with the following:
\begin{itemize}
    \item we train a new set of state-of-the-art deep image classification networks on the same-different tasks;
    \item we probe in more detail already examined models, by removing some key architectural elements to understand to what extent they contribute to the convergence;
    \item we perform an extensive evaluation of the newly trained architectures both in terms of convergence and generalization abilities, reporting training and validation curves.
\end{itemize} 


The rest of the paper is organized as follows:
in Section \ref{sec:related-work}, we review some of the related work in abstract reasoning literature;
in Section \ref{sec:dataset}, we briefly describe the same-different problems from the SVRT dataset; in Section \ref{sec:method}, we present the probed architecture and the training setup;
in Section \ref{sec:experiments}, we discuss our extended experiments; in Section \ref{sec:conclusions} we draw some conclusions on the systematic study performed in this work.

\section{Related Work}
\label{sec:related-work}
\noindent In some studies, abstract and high-level reasoning abilities are probed using some on-purpose generated complex relational tasks. Among the most interesting ones, we find benchmarks such as CLEVR and Sort-of-CLEVR~\citep{Johnson2017CLEVR}, that probed neural networks on the complex R-VQA (Relational Visual Question Answering) task, which consists in answering questions about complex dispositions of simple 3D shapes.
CLEVR is a synthetic dataset composed of 3D rendered scenes, designed on purpose for solving the R-VQA task. Sort-of-CLEVR is a simpler version of CLEVR dealing with 2D shapes and a mix of relational and non-relational questions.

The work in \cite{Santoro2017RelationNetworks} tried to solve the CLEVR R-VQA task by introducing an end-to-end trainable architecture composed of a CNN-based perception module and a novel relation network (RN).
Recently, \cite{messina2018relfeatures,Messina2019RCBIR} extended the relation network model to extract relationship-aware visual features for indexing purposes.


An interesting research direction in relational and abstract understanding is undertaken by ~\cite{Johnson2017} and ~\cite{Mascharka2018}, which developed upon the idea of dynamically assembling an explainable program conditioned on the image-question pair, able to infer the correct answer by performing multiple reasoning steps. They reached more than 99\% accuracy on the CLEVR test set.


Other than the recently developed CLEVR dataset, other benchmarks were introduced to test the relational and abstract reasoning abilities of artificial vision systems.

Some works tackled the abstract reasoning abilities of neural networks by using Raven's Progressive Matrices (RPM). RPMs consist of visual geometric designs with a missing part. The test taker is given a small number of different choices to pick from and fill in the missing piece.
In particular, ~\cite{zhang2019raven} tried to establish a semantic link between vision and reasoning by employing hierarchical representations suitable for relational and analytical thinking. 

Instead, ~\cite{barrett2018abstractreasoning} introduced Procedurally Generated Matrices (PGMs), similar to RPMs but procedurally generated using a detailed algorithm to create a fully controlled environment. They introduced a novel architecture that defeated popular state-of-the-art models like ResNets.


%

In ~\cite{Yang2018WorkingMemory} a synthetic dataset has been introduced to test the abilities of a network of memorizing configurations. Although it is similar in essence to 2D synthetic datasets like Sort-of-CLEVR, it is specifically designed to study the behavior of working memories.

A simple yet powerful dataset was introduced in ~\cite{Fleuret2011}. They introduced the \textit{Synthetic Visual Reasoning Test} (SVRT) dataset, composed of simple images containing closed shapes. It was developed to test the relational and comparison abilities of artificial vision systems.
In ~\cite{Stabinger2016} the authors first showed, using the SVRT dataset, that the tasks involving comparisons between shapes were difficult to solve for convolutional architectures like LeNet and GoogLeNet ~\citep{Szegedy2015GoogLeNet}.

Recently, Messina et al.~\citep{Messina2019SameDifferent} demonstrated that some state-of-the-art deep learning architectures for classifying images, in particular ResNet models, can learn this task, generalizing to some extent.

\section{Review of the Same-different Problems}
\label{sec:dataset}

\noindent In our work, we use the \textit{Synthetic Visual Reasoning Test} (SVRT) benchmark for probing current state-of-the-art deep networks on the same-different tasks. 

SVRT consists of simple 2D images containing simple black closed curves on a white background. Every visual problem in SVRT is divided into two classes: the set of positive examples, which are the images that satisfy the specific rule, and the set of negative examples, which do not satisfy the rule.




In ~\cite{Kim2018NotSoClevr} the authors proposed an exhaustive evaluation of simple CNN-based networks on all the 23 different sub-tasks of the SVRT dataset. According to their findings, the most difficult same-different problems are the ones related to shape comparison under different geometric transformations (problems no. 1, 5, 20, 21). For this reason, in this work, we tackle in great detail these four challenging problems.

In particular, to solve them we are requested to handle the following challenges: \textbf{Problem 1 (P.1)} - detecting the very same shapes, randomly placed in the image, with the same orientation and scale; \textbf{Problem 5 (P.5)} - detecting two pairs of identical shapes, randomly placed in the image. The two images inside every pair have the same orientation and scale; \textbf{Problem 20 (P.20)} - detecting the same shape, translated and flipped along a randomly chosen axis; \textbf{Problem 21 (P.21)} - detecting the same shape, randomly translated, orientated, and scaled.

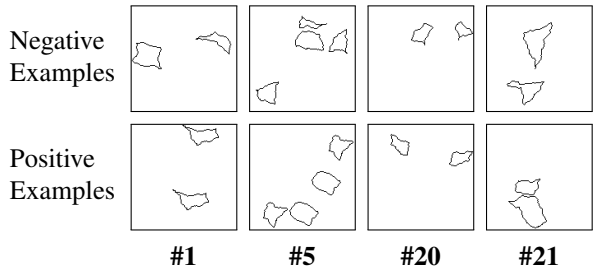
\begin{figure}[t]
\centering
\begin{tikzpicture}[picture format/.style={inner sep=2pt,}]

  \node[picture format]                   (A1)               {\includegraphics[width=0.075\textwidth,frame]{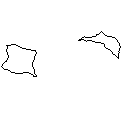}};
  \node[picture format,anchor=north]      (B1) at (A1.south) {\includegraphics[width=0.075\textwidth,frame]{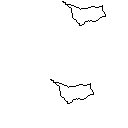}};

  \node[picture format,anchor=north west] (A2) at (A1.north east) {\includegraphics[width=0.075\textwidth,frame]{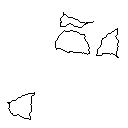}};
  \node[picture format,anchor=north]      (B2) at (A2.south)      {\includegraphics[width=0.075\textwidth,frame]{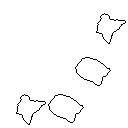}};

  \node[picture format,anchor=north west] (A3) at (A2.north east) {\includegraphics[width=0.075\textwidth,frame]{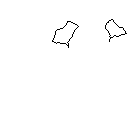}};
  \node[picture format,anchor=north]      (B3) at (A3.south)      {\includegraphics[width=0.075\textwidth,frame]{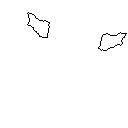}};
  
  \node[picture format,anchor=north west] (A4) at (A3.north east) {\includegraphics[width=0.075\textwidth,frame]{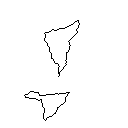}};
  \node[picture format,anchor=north]      (B4) at (A4.south)      {\includegraphics[width=0.075\textwidth,frame]{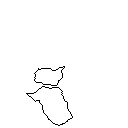}};


  \node[anchor=north] (C1) at (B1.south) {\bfseries \#1};
  \node[anchor=north] (C2) at (B2.south) {\bfseries \#5};
  \node[anchor=north] (C3) at (B3.south) {\bfseries \#20};
  \node[anchor=north] (C4) at (B4.south) {\bfseries \#21};
  
  \node[anchor=east,align=left] (C1) at (A1.west) {Negative\\Examples};
  \node[anchor=east,align=left] (C2) at (B1.west) {Positive\\Examples};

\end{tikzpicture}
\caption{Positive and negative examples from the four SVRT problems.}
\label{fig:svrt-examples}
\end{figure}

Figure~\ref{fig:svrt-examples} shows examples of positive and negative samples for each of the same-different problems under consideration. 

\section{Method}
\label{sec:method}
\noindent This work is aimed at probing state-of-the-art architectures forming the basis of modern computer vision models, and at measuring their abilities to intrinsically perform some abstract reasoning on images.



In our previous work~\citep{Messina2019SameDifferent}, we probed the following state-of-the-art architectures for image classification: AlexNet~\citep{Krizhevsky2012}, VGG-19~\citep{Liu2015Vgg}, three variants of the Resnet~\citep{He2016Resnet}: in order of increasing complexity ResNet-18, ResNet-34 and ResNet-101, and a recently introduced biologically inspired network called CorNet-S~\citep{Kubilius2018CorNet}. 

In this paper, our aim is trying to draw some better conclusions on the architectural factors that contribute to solving the same-different problems. 

For this reason, we probe also two versions of the DenseNet architecture, DenseNet-121 and DenseNet-201, which implement non-residual skip connections, and we explored the Batch Normalized version of the VGG-19.
Furthermore, we try to remove important architectural building-blocks from previously converging architectures to try to isolate the important architectural factors triggering the convergence. In this regard, we try ResNet-18 and ResNet-34 without residual connections, and three variations of the CorNet-S obtained by removing recurrent and/or residual connections.

Following, we describe in more detail the architectures that we will probe on the same-different problems.


\paragraph{VGGs}
VGG-19 contains a simple convolutional architecture comparable with the AlexNet structure, but it is significantly deeper. 
The original VGG-19, however, does not include in its convolutional modules a batch normalization layer.
This could be quite an important detail for reaching network convergence and better stability, especially if the input image is non-normalized.
For this reason, we experimented also with the \textit{VGG-19-BN} architecture, which is the Batch Normalized version of the VGG-19.
It simply includes a \texttt{BatchNorm} layer after each \texttt{Conv2D}, before the ReLU activation.

\paragraph{ResNets}
ResNets introduce residual connections.
This kind of skip connection helps the model to produce incremental differences in the hidden representations, dynamically refining the data passing through the network until it is sufficiently informative for the downstream task. 

The experiments on residual-connection removal from previously converging ResNet-18 and ResNet-34 (namely ResNet-18-WS and ResNet-34-WS, where WS = \textit{Without Skip-connections}) can highlight the role of residual connections in solving the same-different task.

\paragraph{DenseNets}
The novel experiments conducted on DenseNets can spot the differences between a residual network and a network based on generic skip connections. In fact, the DenseNets, differently from ResNets, introduce multiple non-residual skip connections moving lower-level information to each one of the higher-level layers.

\paragraph{CorNet-S}
\cite{Kar2019} introduced this biologically-inspired network which evolves the ResNet architecture by introducing recurrent connections. CorNet-S is inspired by some experimental evidence reported by the authors on primates brain, claiming that the visual cortex could be comprised of recurrent connections.
CorNet-S is designed to mimick four different brain cortical areas involved with vision;
each one of these four blocks is composed of a recurrent connection together with a residual skip connection. 


Thanks to the weight-sharing among timesteps, this architecture has considerably less learnable parameters than ResNets, and it is therefore an overall cheaper architecture.

The introduced architecture CorNet-S-WR, where WR = \textit{Without Recurrent-connections}, is identical to the CorNet-S model but has recurrent connections removed. This modified network is aimed at spotting out what is the influence of recurrent connections as far as the same-different problems are concerned. Similarly, CorNet-S-WS removes the residual connections from the basic CorNet-S model. We try also to remove both the recurrent and the residual links, giving rise to the CorNet-S-WR-WS model.

Note that CorNet-S-WR, when unrolled, constitutes a very shallow network. To conserve the original depth, we stack in sequence the internal modules a number of times equal to the original timesteps proposed by the authors. In this way, CorNet-S-WR is effectively an unrolled version of CorNet-S, with non-shared weights among timesteps.

\begin{figure}[t]
  \centering
  \includegraphics[width=1\linewidth]{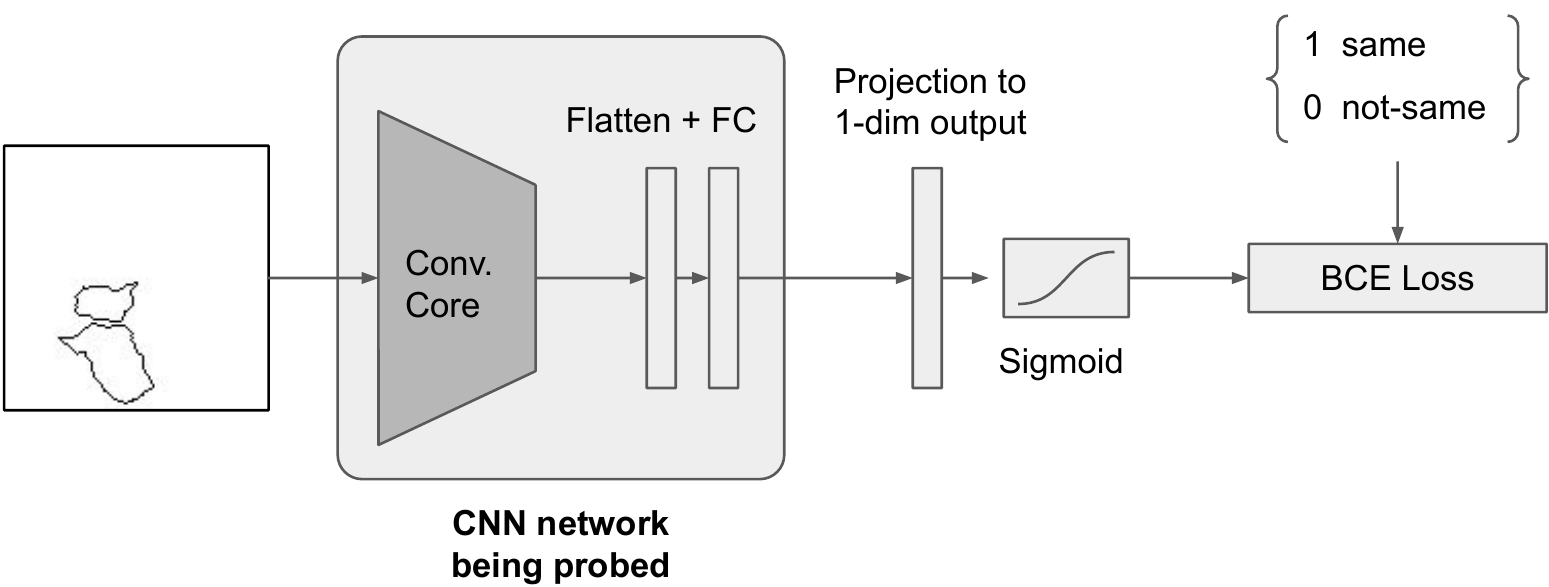}
  \caption{Overview of the network for training on the same-different problems. The architecture of the network in the large light-gray box depends on the specific convolutional network being probed. It is usually composed of a core built of CNN layers plus final FC layers with ReLU activations outputting a fixed-sized vector. We linearly project the output to a single scalar value using a single FC layer. We then normalize this value in the range $[0, 1]$ with a sigmoid activation function before computing the Binary Cross-Entropy (BCE) loss.}
  \label{fig:architecture} 
\end{figure}

\subsection{Training}
All the previously described architectures come with a final classification head since they have been mostly used for the task of image classification. In our scenario, instead, the output must be a binary value indicating if the shapes in the figure correctly satisfy the rules of the specific SVRT problem or not. 

For this reason, we replace the final classification head with a fully-connected layer outputting a single scalar value, normalized in the range $[0, 1]$ using a final sigmoid activation function. The whole network is then trained end-to-end using a binary cross-entropy loss (Figure \ref{fig:architecture}).


\section{Experimental setup}
\label{sec:experiments}

\noindent For each one of the four same-different benchmarks and every model described in Section~\ref{sec:method}, we use 400k images for training, 100k for validation, and 100k for testing. Differently from our previous work where we reported the final accuracy values on the validation set, here we use a separate test set. All the generated images in the SVRT dataset have a size of $128 \times 128$ pixels.

Using 400k training examples could in principle bring to overfitting since the variability of synthetic datasets is often limited. Nevertheless, we claim that being the figures randomly generated in a $128 \times 128$ pixels grid, the probability of generating the same image twice is very low. Furthermore, humans can indeed learn the proposed problems using only a few samples. However, humans use a lot of pre-learned priors to solve these tasks, like the rotation/scale invariance, the concept of shape and closed shape, or the mirroring invariance. Humans, differently from this setup, acquire this knowledge from the experience acquired on a multitude of other tasks, usually by performing transfer learning. 

All the positive and negative examples are perfectly balanced in all the training, validation, and test sets.
For all the probed models, we use SGD as the optimization algorithm, with a momentum of 0.9, weight decay of 1e-4, and a learning rate of 0.1.
We do not use pre-trained weights if they are available. In fact, since these networks are trained on image classification from real-world images, the pre-trained weights cannot deal with the very different distributions of the synthetic SVRT dataset.

The input images are resized to the standard $224 \times 224$ for ResNets and DenseNets architectures.
We perform more experimentation on the VGG-19 and AlexNets models since they are the ones demonstrating major strain.
In particular, for these architectures, we try both to resize the input images to $224 \times 224$ and to keep the original $128 \times 128$ dimensions. 
Furthermore, since AlexNet is designed with no batch-normalization layers, we tried to normalize the input images by subtracting the mean and dividing by the standard deviation of the SVRT dataset.


\newcolumntype{C}{>{\centering\arraybackslash}X}
\newcolumntype{R}{D{,}{\pm}{1.6}}
\newcolumntype{L}{>{\raggedright\arraybackslash}p{5cm}}
\begin{table*}[t]
\begin{center}
\caption{Accuracy values (\%) measured on the test set of the probed architectures, for each of the four SVRT problems. Experiments reaching a perfect chance accuracy are reported in gray. The values reported from~\cite{Stabinger2016, Fleuret2011} did not report any convergence information (CE is n.a.); also, these values are reported with the same number of significant digits as in the original papers.}
\begin{tabular}{Llclclclc}
\toprule
& \multicolumn{2}{c}{\textbf{Problem 1}} & \multicolumn{2}{c}{\textbf{Problem 5}} & \multicolumn{2}{c}{\textbf{Problem 20}} & \multicolumn{2}{c}{\textbf{Problem 21}} \\
\cmidrule(lr){2-3} \cmidrule(lr){4-5} \cmidrule(lr){6-7} \cmidrule(lr){8-9}
\textbf{Model} & \multicolumn{1}{c}{Acc.} & \multicolumn{1}{c}{CE} & 
\multicolumn{1}{c}{Acc.} & \multicolumn{1}{c}{CE} &
\multicolumn{1}{c}{Acc.} & \multicolumn{1}{c}{CE} &
\multicolumn{1}{c}{Acc.} & \multicolumn{1}{c}{CE} \\
\midrule
LeNet \citep{Stabinger2016}                 & 57 & n.a.          & 54 & n.a.        & 55 & n.a.          & 51 & n.a. \\
GoogLeNet \citep{Stabinger2016}          & \textcolor{gray}{50} & n.a.          & \textcolor{gray}{50} & n.a.        & \textcolor{gray}{50} & n.a.          & 51 & n.a. \\
AdaBoost \citep{Fleuret2011}                    & 98 & n.a.          & 87 & n.a.        & 70 & n.a.          & \textcolor{gray}{50} & n.a. \\ 
\midrule
AlexNet              & \textcolor{gray}{\textcolor{gray}{50.0}} & -          & \textcolor{gray}{50.0} & -        & \textcolor{gray}{50.0} & -          & \textcolor{gray}{50.0} & - \\
AlexNet 224x224      & \textcolor{gray}{50.0} & -          & \textcolor{gray}{50.0} & -        & \textcolor{gray}{50.0} & -          & \textcolor{gray}{50.0} & - \\
AlexNet norm.input   & 80.1 & - & \textcolor{gray}{50.0} & - & 76.1 & - & 84.1 & -\\
VGG-19               & \textcolor{gray}{50.0} & -          & \textcolor{gray}{50.0} & -        & \textcolor{gray}{50.0} & -          & \textcolor{gray}{50.0} & - \\
VGG-19 224x224       & \textcolor{gray}{50.0} & -          & \textcolor{gray}{50.0} & -        & \textcolor{gray}{50.0} & -          & \textcolor{gray}{50.0} & - \\
VGG-19-BN            & \textcolor{gray}{50.0} & -          & \textcolor{gray}{50.0} & -        & \textcolor{gray}{50.0} & -          & \textcolor{gray}{50.0} & - \\
VGG-19-BN 224x224    & 93.8	& 1.5 & 93.1 & 6.0 & \textcolor{gray}{50.0} & - & \textcolor{gray}{50.0} & - \\
ResNet-18            & 99.2 & 0.5 & \textbf{99.9} & 2.5 & 95.5 & 2.0 & 96.2 & 17.5 \\
ResNet-18-WS        & 98.9 & 0.5 & 99.5 & 2.0 & 95.7 & 1.0 & 96.7 & 8.5 \\
ResNet-34           & 98.2 & 4.5 & 98.7 & 1.5 & 93.8 & 6.5 & 96.9 & 13.0\\
ResNet-34-WS        & 98.6 & 1.0 & 97.6 & 1.5 & 93.6 & 1.0 & 90.8 & 17.5 \\
ResNet-101          & 99.1 & 3.5 & 96.0 & 3.5 & \textbf{95.8} & 4.0 & 91.1 & 20.5 \\
CorNet-S            & 96.9 & 1.0 & 96.8 & 2.0 & 95.0 & 2.0 & 96.9 & 17.0\\
CorNet-S-WS         & 95.6 & 1.5 & 97.1 & 2.0 & 92.7 & 3.0 & 90.7 & 18.5 \\
CorNet-S-WR         & 94.2 & 1.5 & 91.0 & 7.5 & 91.5 & 4.0 & 88.3 & - \\
CorNet-S-WS-WR      & 93.5 & 1.5 & 92.7 & 8.0 & 91.3 & 7.5 & 86.5 & - \\
DenseNet-121        & \textbf{99.6} & 1.0 & 98.2 & 2.5 & 94.2 & 1.5 & 95.1 & 7.0 \\
DenseNet-201        & 99.5 & 0.5 & 99.3 & 1.5 & 94.3 & 1.5 & \textbf{97.5} & 17.0 \\
\midrule
Human \citep{Fleuret2011}                & 98 &       & 90 &        & 98 &         & 83 &  \\
\bottomrule
\end{tabular}
\label{tab:results-experiment1}
\end{center}
\end{table*}

\subsection{Experiment 1: Convergence}
In this first experiment, we aim to understand if the explored networks can converge on the four same-different problems.
Thus, in this setup, we train the various models, measuring their accuracy on the test set of the same same-different problem.
In this experimental scenario, we are also interested in measuring what is the strain perceived by the network during the training phase. 
For this reason, we desire to capture fine-grained insights during the training phase, for understanding what is the effect of different architectures (or small modifications of the same) on the training curves. Reporting the training curves for all the explored networks is infeasible; however, we can extract some relevant information from the training curves and summarize them under the form of a simple index.

To this aim, similarly to~\citep{Kim2018NotSoClevr}, we extract from the training curves the point (expressed in epochs or fraction of epochs) in which the validation accuracy reaches 90\%. We assume that the more the network is strained, the more examples it needs during the training phase to reach a good accuracy. We call this particular point \textit{convergence epoch} (CE).
%

Together with our measurements, we also report the values as measured by~\cite{Stabinger2016, Fleuret2011} on LeNet, GoogLeNet, and AdaBoost (using feature group 3).

Looking at Table \ref{tab:results-experiment1}, it is clear that most of the configurations derived from AlexNet and VGG-19 architectures are unable to learn or are particularly strained.
More in detail, almost all the VGG-19 configurations remain on the chance level accuracy of 50\%, apart from the VGG-19-BN resized to 224x224. Nevertheless, this configuration can converge on only two out of four problems, and with accuracies far below the state-of-the-art reached with ResNets and DenseNets.
This is the case even for the AlexNet with normalized input images.





Residual networks, as well as DenseNets, are always able to converge obtaining state-of-the-art performances on the proposed tasks.
Also, residual and dense networks defeat humans on three of the four tasks. 

The fact that both residual and dense networks behave so well suggests that there is no significant difference between the residual connections and the DenseNet-like skip connections in this scenario.
Furthermore, the ResNets without recurrent connections, (ResNet-18-WS and ResNet-34-WS) reach almost the same accuracies of the full ResNets, except in the P.21 where ResNet-34-WS obtains a lower accuracy with a higher CE, indicating a little strain with respect to the full ResNet-34. Overall, these results suggest that residual connections may be architectural building-blocks with little impact on the final test accuracy, as far as the convergence is concerned.

\begin{figure*}[t]
  \begin{subfigure}[b]{0.49\textwidth}
      \includegraphics[width=1\linewidth]{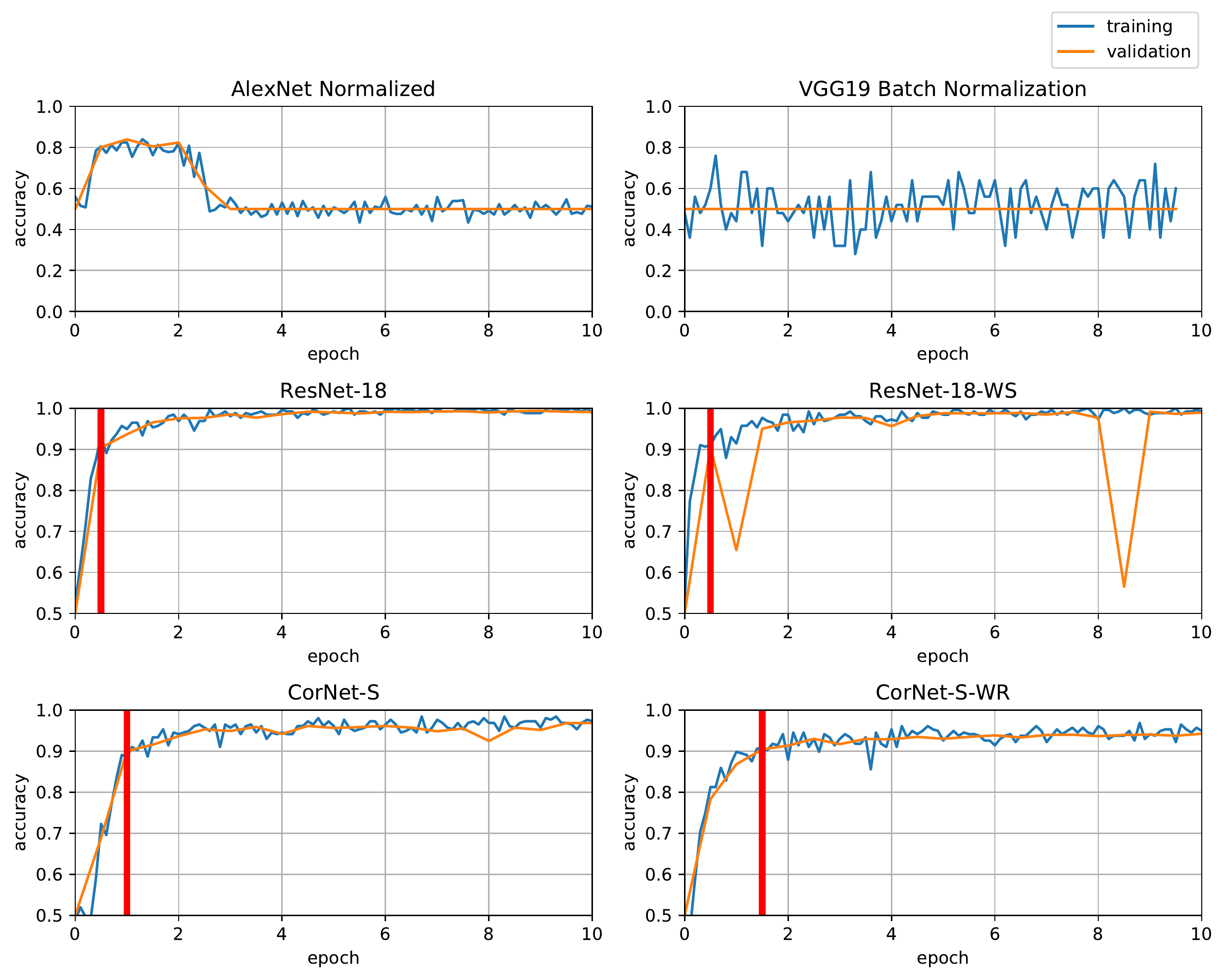}
      \caption{}
      \label{fig:plots_problem1}
  \end{subfigure}
  \rule{1px}{200px}
  \begin{subfigure}[b]{0.49\textwidth}
      \includegraphics[width=1\linewidth]{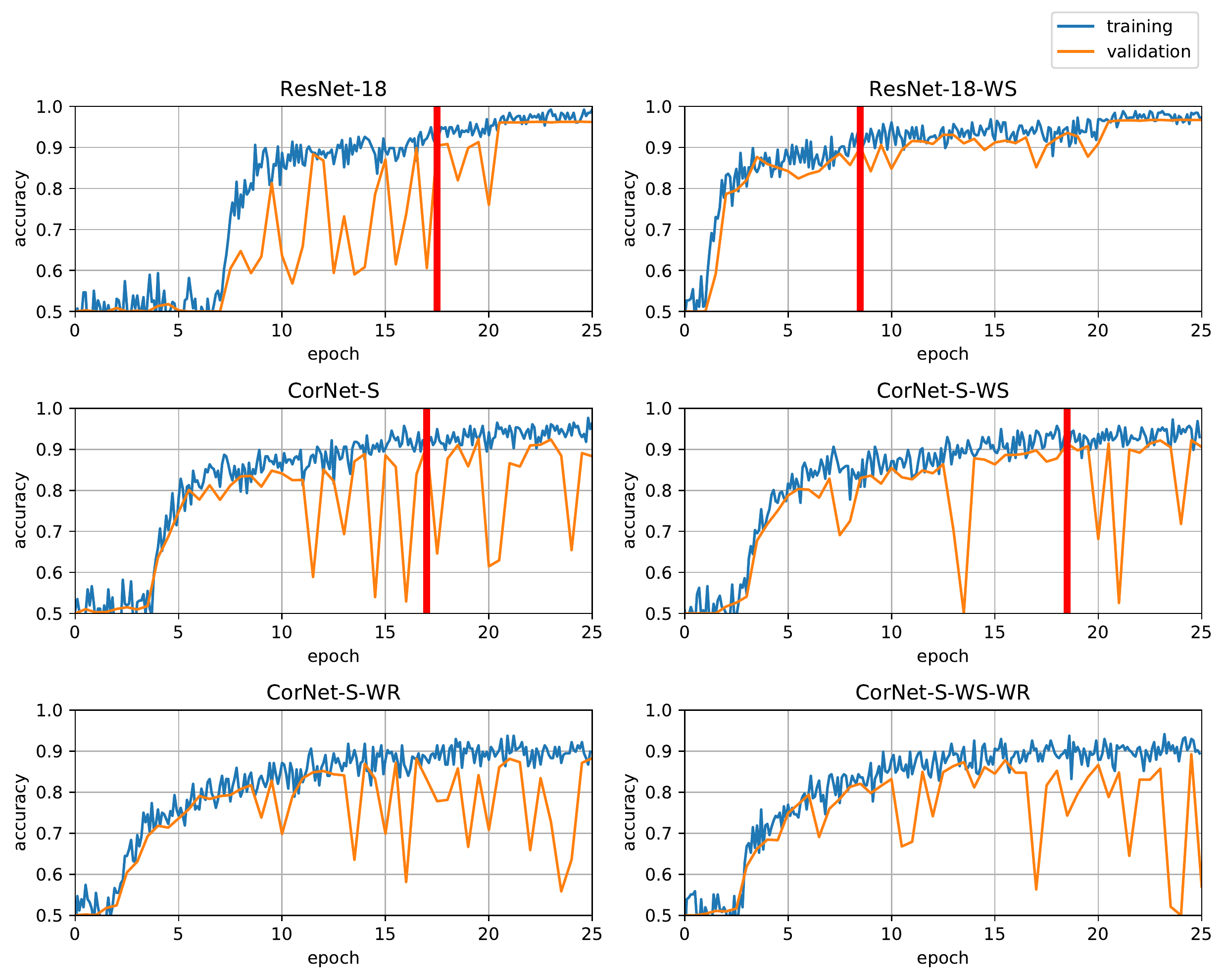}
      \caption{}
      \label{fig:plots_problem21}
  \end{subfigure}
  \caption{Training curves for some of the networks trained on P.1 (a) and P.21 (b). The red vertical line is placed in the correspondence of the convergence epoch (CE). Note that it is not present for the methods that do not reach at least 90\% accuracy on the validation set. For performance reasons we validate the model every half epoch, so we can provide the CE with a resolution of 0.5 epochs.}
\end{figure*}

The comparison among the CorNet-S-WS and CorNet-S-WR shows that the lack of recurrent connections in CorNet-S has a stronger impact than the lack of the residual ones. 
Without recurrent connections, CorNet-S cannot reach 90\% validation accuracy on P.21, leaving the CE for this experiment uncharted. 

Furthermore, the CorNet-S-WS-WR experiment, which lacks both residual and recurrent connections, reaches accuracies on the test set perfectly comparable with the CorNet-S-WR network, suggesting that the removal of residual connections has not a strong impact on the overall convergence accuracies.


We claim that recurrent connections, which implies sharing the weights among timesteps, help in regularizing and stabilizing the network, and they impact considerably on the overall test accuracy when removed.

In Figures \ref{fig:plots_problem1} and \ref{fig:plots_problem21} we report also some detailed training insights for some of the architectures present in Table \ref{tab:results-experiment1} and trained on P.1 and P.21 respectively. In particular, Figure \ref{fig:plots_problem21} shows how P.21 looks immediately more difficult when considering the training curves. The convergence epochs are very noisy and visibly shifted to the right.

It is worth to mention that P.20 is the only problem that, as of now, is still overtaken by humans (last row of Table \ref{tab:results-experiment1}). This can be due to the intrinsic difficulties of deep convolutional neural networks to discern flipped shapes, on which humans instead are very good. It is also interesting to note that the greatest performance gaps among deep networks and humans can be found in P.5 and P.21, probably the most complex problems. Here, humans are heavily defeated by state-of-the-art convolutional neural networks, which probably develop a finer intuition on problems with many degrees of freedom.

\subsection{Experiment 2: Generalization}
Following, we test the generalization abilities of the converged models by measuring their performance on the test set of other problems.

We probe the models trained on P.21 and P.1. In particular, P.21 should force the models to learn most of the required invariances needed to solve all the other problems (translation, scale, rotation). It is interesting to understand if the networks trained on P.21 are also able to solve P.1 and P.5 and to measure their generalization abilities to shapes mirroring (P.20). 

On the other hand, P.1 only requires the networks to learn translation invariance; therefore, it is interesting to understand how well networks trained on this task can deal with rotation, scale, or mirroring invariance (P.20, P.21). It is also worth trying the P.1 generalization abilities to multiple shapes (P.5).

Table \ref{tab:results-experiment2-problem1} reports the accuracies for the most promising models trained on P.1 on the test sets of P.5, P.20, and P.21. Instead, Table \ref{tab:results-experiment2-problem21} provides the accuracies for the most prominent networks trained on P.21 on the test sets of P.1, P.5, P.20.

\subsubsection{Discussion: Training on P.1 and Testing on the others}
Looking at Table \ref{tab:results-experiment2-problem1} it turns out that none of the models obtaining almost-perfect test accuracy on P.1 can understand P.21. This is reasonable since P.21 requires rotation and scale invariances, difficult to acquire for an architecture trained on P.1.

On the other hand, ResNet-34 can generalize quite well to P.5. Note that P.5 requires the models to understand that objects should be clustered into two pairs of possibly identical shapes, and it is not trivial to deduce this information by learning only from single pairs figures.

The architectural changes made to ResNets and CorNet-S have a visible impact on this generalization scenario, especially when recurrent and residual connections are removed. For example, when testing on P.5, ResNet-34 without residual connections (ResNet-34-WS) loses around 10\% with respect to the basic architecture, while CorNet-S-WS and CorNet-S-WR lose 13\% when compared to CorNet-S. The lack of both recurrent and residual connections in this scenario brings to a huge loss in accuracy (18\%). ResNet-18 seems to be an outlier to this trend: on P.20, it obtains a better accuracy when the residual connections are removed.

On P.20, the higher accuracy is reached by CorNet-S. Although approaching a test accuracy far below the optimal one, the clear deviation from chance accuracy suggests that this architecture can partially understand flipped shapes.

Remarkably, the DenseNet networks cannot generalize very well to any of the three test problems.

\newcolumntype{L}{>{\raggedright\arraybackslash}p{2.9cm}}
\begin{table}[t]
\begin{center}
\caption{Accuracy values (\%) measured on the probed architectures, by training on P.1 and testing on the test sets of the other three problems.}
\begin{tabular}{Lccc}
\toprule
\textbf{Model} & \textbf{Test P.5} & \textbf{Test P.20} & \textbf{Test P.21} \\
\midrule
ResNet-18               & 56.5 & 55.6 & 51.6 \\
ResNet-18-WS           & 56.4 & 58.4 & 51.2 \\
ResNet-34               & \textbf{84.4} & 61.6 & 51.5\\
ResNet-34-WS            & 75.4 & 61.3 & 51.5\\
\midrule
CorNet-S                & 73.6 & \textbf{78.7} & 52.0\\
CorNet-S-WS             & 64.6 & 76.8 & 51.7\\
CorNet-S-WR             & 63.9 & 71.3 & \textbf{52.5}\\
CorNet-S-WS-WR          & 60.7 & 76.2 & 52.4\\
\midrule
DenseNet-121            & 58.8 & 55.3 & 51.2 \\
DenseNet-201            & 56.2 & 54.5 & 51.3 \\
\bottomrule
\end{tabular}
\label{tab:results-experiment2-problem1}
\end{center}
\end{table}

\subsubsection{Discussion: Training on P.21 and Testing on the others}
Table \ref{tab:results-experiment2-problem21} shows how the models trained on P.21 can understand also P.1. This is expected since P.1 is a subset of P.21 that does not deal with scales and orientations of the shapes.

A remarkable result can be observed on the networks tested on P.20: the great part of the networks trained on P.21 can solve this problem almost perfectly, although P.21 does not carry the concept of shape mirroring.
In particular, DenseNet-201 can reach state-of-the-art results on both P.1 and P.20.

In this generalization scenario, CorNet-S and ResNets suffer from the removal of residual and recurrent connections only when addressing P.1 and P.20, reaching the minimum accuracy on these problems with the CorNet-S-WS-WR, where both the residual and recurrent links are missing. ResNet-18-WS defines an exception to this trend since it can perform better than the full ResNet-18 on these two problems.

However, there is an interesting trend when comparing P.1 or P.20, with P.5. If we concentrate on the various CorNet-S versions, we notice that there is a decreasing accuracy trend on P.1 and P.20 when the recurrent and skip connections are gradually removed. Instead, an increasing trend is visible for P.5, although the absolute values for P.5 remain very low. The same thing happens for ResNets and DenseNets.

The low absolute accuracy values obtained in P.5 suggest that it is not sufficient to be invariant to rotation, scale, or translation to understand this problem. 
In general, zero-shot generalization to P.5 is difficult: it is needed to observe a few samples of the target problem to understand the new rules. 

\begin{table}[t]
\begin{center}
\caption{Accuracy values measured on the probed architectures, by training on P.21 and testing on the test sets of the other three problems.}
\begin{tabular}{Lccc}
\toprule
\textbf{Model} & \textbf{Test P.1} & \textbf{Test P.5} & \textbf{Test P.20} \\
\midrule
ResNet-18               & 97.9 & 54.2 & 96.0 \\
ResNet-18-WS           & 98.3 & 53.3 & 96.6 \\
ResNet-34               & 98.3 & 59.4 & 96.6\\
ResNet-34-WS            & 94.2 & \textbf{63.4} & 91.7\\
\midrule
CorNet-S                & 98.6 & 54.2 & 97.0\\
CorNet-S-WS             & 95.6 & 59.1 & 91.7\\
CorNet-S-WR             & 92.4 & 61.4 & 89.9\\
CorNet-S-WS-WR          & 91.7 & 62.4 & 87.9\\
\midrule
DenseNet-121            & 96.9 & 55.7 & 95.1 \\
DenseNet-201            & \textbf{98.9} & 50.8 & \textbf{97.4} \\
\bottomrule
\end{tabular}
\label{tab:results-experiment2-problem21}
\end{center}
\end{table}

%



\section{Conclusions}
\label{sec:conclusions}
\noindent In this work, we analyzed to what extent very-deep convolutional neural networks can deal with the \textit{same-different} challenging tasks.


Considering the SVRT visual challenge, our results show that ResNets, DenseNets, and CorNet-S (a biologically-inspired architecture similar to ResNet architecture) can correctly understand the challenging images and generalize to never seen shapes. We found that all these architectures found P.21 the harder to learn.
Older models such as AlexNet and VGG cannot in many cases converge on the proposed problems. 

In this paper, we conducted additional experiments with respect to our previous work~\citep{Messina2019SameDifferent}, that demonstrated how residual connections are not the only significant architectural details for solving the same-different task. 
Also, it seems that residual and recurrent connections have important roles in the zero-shot generalization to similar problems. 

In the end, we showed that recent convolutional networks for image classification can reach state-of-the-art results on all these challenging same-different problems from the SVRT dataset. Also, we found that they can reach super-human performances on P.1, P.5, and P.21.


We think that the development of the abstract and relational abilities of neural networks is an important leap towards achieving some interesting new tasks, such as aesthetic judgment in images and even in music, with wide applications in many fields (e.g., cultural heritage preservation).

\section*{Acknowledgments}
\noindent This work was partially supported by “Intelligenza Artificiale per il Monitoraggio Visuale
dei Siti Culturali" (AI4CHSites) CNR4C program, CUP B15J19001040004 and by the AI4EU project, funded by the EC (H2020 - Contract n. 825619).

\bibliographystyle{model2-names}
\bibliography{refs}

\end{document}